# Adaptive Regularization for Weight Matrices

**Koby Crammer** KOBY@EE.TECHNION.AC.IL
Department of Electrical Engineering, Technion, Haifa 32000, Israel

**Gal Chechik** GAL.CHECHIK@BIU.AC.IL
The Gonda Brain Research Center, Bar Ilan University, and Google research

## Abstract

Algorithms for learning distributions over weight-vectors, such as AROW (Crammer et al., 2009) were recently shown empirically to achieve state-of-the-art performance at various problems, with strong theoretical guaranties. Extending these algorithms to matrix models pose challenges since the number of free parameters in the covariance of the distribution scales as $n^4$ with the dimension $n$ of the matrix, and $n$ tends to be large in real applications. We describe, analyze and experiment with two new algorithms for learning distribution of matrix models. Our first algorithm maintains a diagonal covariance over the parameters and can handle large covariance matrices. The second algorithm factors the covariance to capture inter-features correlation while keeping the number of parameters linear in the size of the original matrix. We analyze both algorithms in the mistake bound model and show a superior precision performance of our approach over other algorithms in two tasks: retrieving similar images, and ranking similar documents. The factored algorithm is shown to attain faster convergence rate.

## 1. Introduction

Many machine learning tasks involve models in the form of a matrix. As an important example, consider the problem of linear metric learning where the dissimilarity between a pair of samples is measured using the Mahalanobis distance, parametrized by a positive semi-definite matrix. A second important example is the matrix model obtained when learning multiple linear classifiers regularized jointly, like in the case of object recognition with many classes.

 

Many algorithms were developed for learning these two tasks, including online algorithms developed recently in the context of classification and ranking costs (Davis et al., 2007; Jain et al., 2008; Chechik et al., 2009).

While such linear matrix models are common for metric and multiclass learning, the broader class of ”vector” linear model are a popular choice in many domains since they provide a good balance between simplicity, scalability and performance. Methods to generate linear classifiers from data have flourished in the past decade, including SVMImportantly, when learning linear models, it was recently shown that modeling the second order information about the set of models (Crammer et al. (2009) and the references therein), or using this information during training (Duchi et al., 2010) improves the convergence rate of the learning algorithms as well as the performance of the resulting classifiers. These very effective methods were developed primarily for handling vector models, and were not designed to handle matrix models.

At first sight, problems that involve learning matrices could be handled directly using methods developed for learning vectors, including the second order methods described above. In practice however, matrix models often pose a challenge to scalability, since both their memory and their runtime complexity scale quadratically with the data dimensionality $n$. Modeling second order interactions between features may therefore require $n^4$ parameters, limiting these methods to relatively low dimensional data.

In this paper we study second-order methods for learning matrix models and test them in the context of similarity learning. We describe AROMA (Adaptive Regularization Of MAtrix models) an online algorithm that learns a distribution of matrix models. Since maintaining a full covariance matrix over the parameters would not be feasible for large dimensions, we describe models that capture part of the covariance structure. We first describe a simple model with a diagonal covariance matrix. While this model scales well to large matrices, it fails to model correlations between features which could be crucial in some applications. We further describe a factored model which is still linear in the number of parameters (quadratic in the dimension), yet captures some of the correlations between features.

In the context of metric and similarity learning, AROMA can be used to learn a distribution over metrics, instead of a single metric. We evaluate AROMA in two tasks of retrieving images and documents by evaluating similarity between objects. We find that the two AROMA variants outperform competing methods by a large gap. Additionally, the more involved variant convergence faster than all other methods evaluated. As far as we know, this makes it the state-of-the-art method for the extensively studied task of linear similarity learning.

**Notation:** In this work we often consider the bilinear form $\mathbf{q}^\top W \mathbf{p}$ where $\mathbf{q} \in \mathbb{R}^m$, $\mathbf{p} \in \mathbb{R}^n$ and $W \in \mathbb{R}^{m \times n}$. Given such a matrix $W$, we denote by $\mathbf{vec}(W) \in \mathbb{R}^{mn}$ the vector generated by "stacking" the columns of the matrix $W$. Using this operator we can write the bilinear form as an inner product $\mathbf{q}^\top W \mathbf{p} = \mathbf{vec}(W) \cdot \mathbf{vec}\left(\mathbf{p}\mathbf{q}^\top\right)$. We denote by $\boldsymbol{x} \odot \boldsymbol{z}$ the element-wise product of two vectors (or matrices) and by $sum(A)$ the sum of the elements of the matrix or vector $A$. We denote by $|\boldsymbol{x}|_0$ to be the number of non-zero elements of the vector $\boldsymbol{x}$, known as $\ell_0$ norm.

Given two square matrices $\Lambda \in \mathbb{R}^{m \times m}$ and $\Omega \in \mathbb{R}^{n \times n}$ we denote their Kronecker product by $\Lambda \otimes \Omega$. This is a matrix of size $mn \times mn$ that is composed of blocks, where the $(i,j)^{th}$ block is $\Lambda_{i,j}\Omega$. Finally, (Sx) refers to the equation x in a longer version of this manuscript provided online[1].

## 2. Problem Setting

We focus on the problem of learning a linear similarity measure between pairs of objects $\mathbf{q} \in \mathbb{R}^m$, $\mathbf{p} \in \mathbb{R}^n$, in the form of $S_W(\mathbf{q}, \mathbf{p}) = \mathbf{q}^\top W \mathbf{p}$. This similarity measure is related to metric learning models of the form $(\mathbf{q} - \mathbf{p})^\top W (\mathbf{q} - \mathbf{p})$ for square matrices $W$, and becomes equivalent to it when all vectors $\mathbf{p}$ and $\mathbf{q}$ have a constant $W$-norm. Interestingly, the similarity measure $S_W(\mathbf{q}, \mathbf{p})$ does not have to be symmetric, and may even be defined for objects from with different dimensions $m \neq n$ (non-square $W$). In general, it allows to learn a measure of relatedness between objects from different domains, like images and sounds or images and text (as in Grangier & Bengio, 2008). Importantly, when the vectors representing both query and object are sparse and contain only few elements, $|\mathbf{q}|_0 = k_\mathbf{q}$, $|\mathbf{p}|_0 = k_\mathbf{p}$ computing the similarity score takes only $k_\mathbf{q} k_\mathbf{p}$ operations instead of $mn$ for dense vectors.

We address a weak-supervision setup where training is based on *relative similarity*. Here, we are allowed to sample triplets of objects, each triplet containing a "query object" $\mathbf{q} \in \mathbb{R}^m$ and two candidate objects $\mathbf{p}^+, \mathbf{p}^- \in \mathbb{R}^n$, where it is known that the object $\mathbf{p}^+$ is more related (or similar) to the query $\mathbf{q}$ than the other object $\mathbf{p}^-$.

Importantly, the relative similarity learning setup does not assume that there exists an *absolute* numerical level of similarity between an object and a query, or that the learner has access to it. Training therefore assumes a weaker type of supervision, making it easier to collect labeled data either from human raters, or by collecting indirect data about association of object pairs. For example, two web pages can be ranked by their similarity to a third web page by the number of users visiting them within the same session.

Formally, our goal is to learn a bi-linear similarity scoring function $\mathcal{S}_W(\mathbf{q}, \mathbf{p}) = \mathbf{q}^\top W \mathbf{p}$ parametrized by $W \in \mathbb{R}^{m \times n}$ such that the total ordering induced by the similarity function over objects $\mathbf{p}$ would be consistent with the partial ordering information given about $\mathbf{p}^-$ and a query $\mathbf{q}$. A similar model was recently studied in different contexts (McFee & Lanckriet, 2012; Kulis et al., 2011; Weston et al., 2011).

We formalize training as a constrained optimization problem and require that this relation between the induced ranking and the partial information of ordering holds with a safety margin,

$$\mathcal{S}_W(\mathbf{q}, \mathbf{p}^+) \geq \mathcal{S}_W(\mathbf{q}, \mathbf{p}^-) + 1 \; . \tag{1}$$

More specifically, we develop an *online* algorithm that allows to rank objects by their similarity to a "query object" $\mathbf{q}$. Like online prediction algorithms, online retrieval algorithms work in rounds. On round $i$, the algorithm receives a triplet composed of a query $\mathbf{q}_i \in \mathbb{R}^m$ and two possible outcomes $\mathbf{p}_i^+, \mathbf{p}_i^- \in \mathbb{R}^n$. The algorithm than outputs a single bit indicating which outcome is better for the given query. It then receives the correct answer and updates its model.

To learn a scoring function that obeys (1), we define a hinge loss over the triplet $(\mathbf{q}, \mathbf{p}^+, \mathbf{p}^-)$

$$\ell_W(\mathbf{q},\mathbf{p}^+,\mathbf{p}^-) \quad = \max\left(0, 1 - \mathbf{q}^\top W (\mathbf{p}^+ - \mathbf{p}^-)\right) \; . \tag{2}$$

In what follows, we describe two online algorithms to minimize this loss while modeling the distribution of matrix models $W$. We first review previous work on learning such distributions for vector models.

## 3. Adaptive Regularization of Weights

We first describe the AROW algorithm that was designed for binary classification of vector inputs $\boldsymbol{x} \in \mathbb{R}^d$ and introduced by Crammer et al. (2009).

The key idea of AROW (Dredze et al., 2008, and its predecessors), is that instead of maintaining a single vector $\boldsymbol{w}$ during learning, AROW maintains a *distribution* over possible models. Specifically, AROW maintains a Gaussian distribution over vectors denoted by $\mathcal{N}(\boldsymbol{w}, \Sigma)$, where $\boldsymbol{w} \in \mathbb{R}^d$ and $\Sigma \in \mathbb{R}^{d\times d}$. The mean $\boldsymbol{w}$ encodes the knowledge of the algorithm about the weight features (linear model), and is used to make predictions. The covariance $\Sigma$ captures the notion of confidence in the weights, and is used during training to set an effective learning rate for features with different statistics. AROW was motivated by tasks in natural language processing, where many features are very rare and a few features are frequent.

[1] webee.technion.ac.il/people/koby/publications/aroma_icml12long.pdf

AROW is an online algorithm that works in rounds. On the $i$-th round, the algorithm receives an input $\boldsymbol{x}_i \in \mathbb{R}^d$ and employs its current model to make a prediction $\hat{y}_i \in \{\pm 1\}$. It then receives the true label $y_i \in \{\pm 1\}$ and suffers a loss $\ell(y_i, \hat{y}_i)$. Finally, the algorithm updates its prediction rule using the pair $(\boldsymbol{x}_i, y_i)$ and proceeds to the next round.

AROW updates its current model parameters $\boldsymbol{w}$ and $\Sigma$ by minimizing the following objective function

$$\begin{aligned}\mathcal{L}_{AROW} &= \mathrm{D}_{\mathrm{KL}}\left(\mathcal{N}(\boldsymbol{w}, \Sigma) \,\|\, \mathcal{N}(\boldsymbol{w}_{t-1}, \Sigma_{t-1})\right) \quad (3)\\ &+\tfrac{1}{2r}\Big(\max\{0, 1 - y_i \boldsymbol{x}_i^\top \boldsymbol{w}\}\Big)^2 + \tfrac{1}{2r}\boldsymbol{x}_i^\top \Sigma \boldsymbol{x}_i\ ,\end{aligned}$$

where $\mathrm{D}_{\mathrm{KL}}$ is the Kullback-Leibler divergence. This objective aims to find a model that classifies the sample $(\boldsymbol{x}_i, y_i)$ correctly, while keeping the distribution from changing abruptly at a single iteration.

The minimum of the objective in Eq. (3) was shown by Crammer et al. (2009) to be obtained by the update rule:

$$\begin{aligned}\boldsymbol{w}_i &= \boldsymbol{w}_{i-1} + \frac{\max\left(0, 1 - y_i \boldsymbol{x}_i^\top \boldsymbol{w}_{i-1}\right)}{\boldsymbol{x}_i^\top \Sigma_{i-1}\boldsymbol{x}_i + r}\Sigma_{i-1} y_i \boldsymbol{x}_i, \quad (4)\\ \Sigma_i &= \Sigma_{i-1} - \frac{\Sigma_{i-1}\boldsymbol{x}_i \boldsymbol{x}_i^\top \Sigma_{i-1}}{r + \boldsymbol{x}_i^\top \Sigma_{i-1}\boldsymbol{x}_i} \quad .\end{aligned}$$

AROW was shown to attain state-of-the-art performance on many problems (Crammer et al., 2009; Duchi et al., 2010) and its performance is analyzed both for full covariance matrices (Crammer et al., 2009) and diagonal covariance matrices (Orabona & Crammer, 2010). In the next section, and in this entire paper, we lift AROW to matrices, while maintaining both memory and speed efficiency.

# 4. Modeling Uncertainty over Matrices

As with online classification learning, online retrieval algorithms work in rounds. At round $i$ the algorithm receives a triplet composed of a query $\mathbf{q}_i \in \mathbb{R}^m$ and two possible outcomes $\mathbf{p}_i^+, \mathbf{p}_i^- \in \mathbb{R}^n$. The algorithm than outputs a single bit indicating which outcome is better for the given query. It then receives the correct answer and updates its model. For simplicity, we assume that the first outcome is always preferable, namely, given $\mathbf{q}_i$ the algorithm should rank $\mathbf{p}_i^+$ over $\mathbf{p}_i^-$. We now consider the problem of modeling uncertainty over matrices, in the context of online-learning similarity measures that obeys (1), and describe algorithms to minimize the loss in (2).

A naive approach to model uncertainty over matrices would be to to use the linearity of the ranking function $\mathcal{S}_W(\mathbf{q}, \mathbf{p})$ in $W$, and write S as an inner product between two vectors $\mathbf{q}^\top W \mathbf{p} = \mathbf{vec}(W) \cdot \mathbf{vec}\left(\mathbf{p}\mathbf{q}^\top\right)$. Here, learning over matrices of dimension $m \times n$ is viewed simply as learning over vectors of dimension $1 \times mn$. After transforming the matrix model into a vector, then the original AROW algorithm for vectors can be applied.

Unfortunately, this approach requires to maintain the mean parameters as a vector of of size $mn$ and the full covariance matrix of size $(mn) \times (mn)$. Even for moderate dimension values of $m$ and $n$, the size of a full covariance matrix $m^2 n^2$ cannot be stored in memory. For instance, with $m = n = 10^3$, the dimension of the vectorized model is $mn = 10^6$ and the full covariance matrix requires $10^{12}$ parameters. Designing second order algorithms for matrices thus requires to model the covariance in a more compact way. We now discuss and develop two such compact representations and learning algorithms: a diagonal covariance, and a factorized covariance.

## 4.1. Diagonal Covariance

Our first algorithm restricts the covariance matrices to be diagonal, using only $mn$ non-zero elements (the size of the similarity measure $W$). Denote by $\boldsymbol{\sigma} \in \mathbb{R}^{mn}$ the diagonal elements of the covariance matrix. The update (4) becomes

$$\boldsymbol{w}_i = \boldsymbol{w}_{i-1} + \frac{\max\left(0, 1 - y_i \boldsymbol{x}_i^\top \boldsymbol{w}_{i-1}\right)}{sum(\boldsymbol{x}_i^\top \odot \boldsymbol{\sigma}_{i-1} \odot \boldsymbol{x}_i) + r} y_i \boldsymbol{\sigma}_{i-1} \odot \boldsymbol{x}_i$$

and the covariance is,

$$\boldsymbol{\sigma}_i = \boldsymbol{\sigma}_{i-1} - \frac{\boldsymbol{\sigma}_{i-1} \odot \boldsymbol{x}_i \odot \boldsymbol{x}_i \odot \boldsymbol{\sigma}_{i-1}}{r + sum(\boldsymbol{x}_i^\top \odot \boldsymbol{\sigma}_{i-1} \odot \boldsymbol{x}_i)}\ .$$

We denote by $\Sigma \in \mathbb{R}^{m\times n}$ the covariance matrix that maintains one element per feature, and thus is diagonal-like, although it is rectangular in shape. We identify $\boldsymbol{x}_i = \mathbf{q}_i \mathbf{p}_i^\top$, $\mathbf{p}_i = \mathbf{p}_i^+ - \mathbf{p}_i^-$ and $y_i = 1$, to get the update in the notation used for matrix-similarity measures,

$$\begin{aligned}W_i &= W_{i-1} + \alpha_i \Sigma_{i-1} \odot \left(\mathbf{q}_i \mathbf{p}_i^\top\right)\\ \text{where } \alpha_i &= \frac{\max\left(0, 1 - \mathbf{q}_i^\top W_{i-1}\mathbf{p}_i\right)}{sum(\mathbf{q}_i\mathbf{p}_i^\top \odot \Sigma_{i-1} \odot \mathbf{q}_i\mathbf{p}_i^\top) + r} \quad (5)\end{aligned}$$

$$\text{and} \quad \Sigma_i = \Sigma_{i-1} - \frac{\Sigma_{i-1} \odot \mathbf{q}_i\mathbf{p}_i^\top \odot \mathbf{q}_i\mathbf{p}_i^\top \odot \Sigma_{i-1}}{sum(\mathbf{q}_i\mathbf{p}_i^\top \odot \Sigma_{i-1} \odot \mathbf{q}_i\mathbf{p}_i^\top) + r}\ . \quad (6)$$

**Algorithm 1: diagonal-AROMA**

**Input parameters** A scalar $r$

**Initialize** $W_0 = 0 \in \mathbb{R}^{m\times n}$ , $\Sigma_0 = \mathbf{1} \in \mathbb{R}^{m\times n}$
**For** $i = 1, \ldots, N$

- Sample a query $\mathbf{q}_i \in \mathbb{R}^m$ and two images $\mathbf{p}_i^+, \mathbf{p}_i^- \in \mathbb{R}^n$, such that $\mathbf{p}_i^+$ should be ranked above $\mathbf{p}_i^-$
- Define $\mathbf{p}_i = \mathbf{p}_i^+ - \mathbf{p}_i^-$
- If $1 > \mathbf{q}_i^\top W_{i-1}\mathbf{p}_i$ then update:
  - Update $W_i = W_{i-1} + \alpha_i \Sigma_{i-1} \odot \left(\mathbf{q}_i \mathbf{p}_i^\top\right)$ where $$\alpha_i = \frac{\max\left(0, 1-\mathbf{q}_i^\top W_{i-1}\mathbf{p}_i\right)}{sum(\mathbf{q}_i\mathbf{p}_i^\top \odot \Sigma_{i-1} \odot \mathbf{q}_i\mathbf{p}_i^\top) + r} \quad (5)$$
  - Update $\Sigma_i = \Sigma_{i-1} - \frac{\Sigma_{i-1}\odot \mathbf{q}_i\mathbf{p}_i^\top \odot \mathbf{q}_i\mathbf{p}_i^\top \Sigma_{i-1}}{sum(\mathbf{q}_i\mathbf{p}_i^\top \odot \Sigma_{i-1} \odot \mathbf{q}_i\mathbf{p}_i^\top) + r}$ (6)

**Output:** A weight matrix $W_N$ and its confidence $\Sigma_N$

*Figure 1.* The d-AROMA algorithm for similarity measures.

We call the algorithm d(iagonal)-AROMA for *diagonal-Adaptive Regularization Of MAtrix models*, and it is summarized in Fig. 1. The memory required for d-AROMA is $\Theta(mn)$ - the space needed to store both $W$ and $\Sigma$. The time complexity is $\Theta(mn)$ as all operations involve component-wise operations between $W$ and $\Sigma$; and $\mathbf{p}$ and $\mathbf{q}$.

Before proceeding to describe the next algorithm we state a mistake bound for d-AROMA. Let $\mathcal{M}$ be the set of rounds for which the algorithm made a prediction mistake and let $\mathcal{U}$ be the set of example indices for which the algorithm made an update, yet no mistake occurred. Then,

**Theorem 1** *Let $V$ be any similarity matrix. Assume the algorithm is executed on any sequence then the total no. of mistakes it performs is bounded by,*

$$|\mathcal{M}| \le \sum_{i\in\mathcal{M}\cup\mathcal{U}} \max\left\{0, 1-\mathbf{q}_i^\top V \mathbf{p}_i\right\} - |\mathcal{U}| + \sqrt{\|V\|_{Fro}^2 + \frac{1}{r}\sum_{k=1,l=1}^{m,n} V_{k,l}^2 \sum_{i\mathcal{M}\cup\mathcal{U}} \mathbf{q}_{i,k}^2 \mathbf{p}_{i,l}^2} \times \sqrt{r\sum_{k=1,l=1}^{m,n} \log\left(\frac{\sum_{i\mathcal{M}\cup\mathcal{U}} \mathbf{q}_{i,k}^2\mathbf{p}_{i,l}^2}{r} + 1\right) + 2|\mathcal{U}|} \, .$$

The proof is omitted due to lack of space and is similar in spirit to the analysis in section 4.3 of Orabona & Crammer (2010). As in their, analysis we expect the bound to be small if either the combination of the $k^{th}$ feature of the query $\mathbf{q}_{i,k}$ and of the $l^{th}$ feature of the output difference $\mathbf{p}_{i,l}$ is rare (that is $\sum_{i\in\mathcal{M}\cup\mathcal{U}} \mathbf{q}_{i,k}^2\mathbf{p}_{i,l}^2$ is small), or that this combination is not useful for prediction, that is, $V_{k,l}^2$ is small. When most feature combinations fall under one of these two cases, we expect the second term in the first square-root term to be small and most of the values of the log function to be close to zero. Unlike the vector-variant of this analysis, here it is not required that the input features are sparse. Instead, we only require that for some inputs the query is sparse and for other inputs the difference between the objects is sparse, but not necessarily both.

### 4.2. Factored Covariance

Our second approach to model the distribution of similarity matrices is based on factorizing the covariance matrix in a way that captures *separately* correlations in the "input" (right side of the similarity matrix) and in the "output" (left side). To describe our second algorithm, we use the definition of a *matrix-variate normal distribution* (Gupta & Nagar, 1999).

**Definition 1** *A random matrix $X \in \mathbb{R}^{m\times n}$ is said to have a matrix variate normal distribution with mean matrix $W \in \mathbb{R}^{m\times n}$ and covariance matrix $\Omega \otimes \Lambda$ where $\Lambda \in \mathbb{R}^{m\times m}$ and $\Omega \in \mathbb{R}^{n\times n}$ are both symmetric and PSD, if* $\mathbf{vec}(X) \sim \mathcal{N}(\mathbf{vec}(W), \Omega\otimes\Lambda)$. *Matrix variate normal distributions are denoted by $\mathcal{N}(W, \Omega\otimes\Lambda)$.*

Gupta & Nagar (1999) show (Thm. 2.2.1) that the probability density of a matrix variate normal distribution is,

$$p(X|W,\Omega,\Lambda) = (2\pi)^{-\frac{1}{2}mn} \det(\Lambda)^{-\frac{1}{2}n} \det(\Omega)^{-\frac{1}{2}m} \times \exp\left\{-\frac{1}{2}\mathrm{Tr}\left(\Lambda^{-1}(X-W)\Omega^{-1}(X-W)^\top\right)\right\} . \quad (7)$$

We derive our algorithm by revisiting the objective of AROW (3) and compute the three terms of that objective for our model. For the first term, we use (7) and obtain that the KL divergence between two matrix-variate normal distributions is (up to additive constants),

$$\begin{aligned} &\mathrm{D_{KL}}\left(\mathcal{N}(W,\Omega\otimes\Lambda) \,\|\, \mathcal{N}(W_{i-1},\Omega_{i-1}\otimes\Lambda_{i-1})\right) \quad (8)\\ &= \frac{1}{2}n\log\left(\frac{\det\Lambda_{i-1}}{\det\Lambda}\right) + \frac{1}{2}m\log\left(\frac{\det\Omega_{i-1}}{\det\Omega}\right)\\ &\quad+\frac{1}{2}\mathrm{Tr}\left(\Lambda_{i-1}^{-1}\Lambda\right)\mathrm{Tr}\left(\Omega_{i-1}^{-1}\Omega\right)\\ &\quad+\frac{1}{2}\mathrm{Tr}\left(\Lambda_{i-1}^{-1}(W-W_{i-1})\Omega_{i-1}^{-1}(W-W_{i-1})^\top\right) . \end{aligned}$$

For the second term of (3), we use $\mathbf{q}^\top W\mathbf{p} = \mathbf{vec}(W)\cdot\mathbf{vec}\left(\mathbf{p}\mathbf{q}^\top\right)$ as discussed above, to compute

$$\left(\max\left\{0, 1-\mathbf{q}^\top W\mathbf{p}\right\}\right)^2 . \quad (9)$$

Finally, the third term is,

$$\begin{aligned} \mathbf{vec}\left(\mathbf{p}\mathbf{q}^\top\right)^\top (\Lambda\otimes\Omega)\mathbf{vec}\left(\mathbf{p}\mathbf{q}^\top\right) &= \mathbf{vec}\left(\mathbf{p}\mathbf{q}^\top\right)^\top \mathbf{vec}\left(\Omega\mathbf{p}\mathbf{q}^\top\Lambda\right)\\ &= \left(\mathbf{p}^\top\Omega\mathbf{p}\right)\left(\mathbf{q}^\top\Lambda\mathbf{q}\right) , \quad (10) \end{aligned}$$

where we used the identities $\textbf{vec}\left(AXC\right) = \left(C^\top \otimes A\right)\textbf{vec}\left(X\right)$ and $\textbf{vec}\left(A^\top\right)^\top \textbf{vec}\left(C\right) = \text{Tr}\left(AC\right)$. Combining (8), (9) and (10) we get the optimization problem describing the update of the algorithm,

$$\begin{aligned} &\frac{1}{2}n\log\left(\frac{\det \Lambda_{i-1}}{\det \Lambda}\right) + \frac{1}{2}m\log\left(\frac{\det \Omega_{i-1}}{\det \Omega}\right) \\ &+ \frac{1}{2}\text{Tr}\left(\Lambda_{i-1}^{-1}\left(W - W_{i-1}\right)\Omega_{i-1}^{-1}\left(W - W_{i-1}\right)^\top\right) \\ &+ \frac{1}{2r}\left(\max\left\{0, 1-\mathbf{q}^\top W\mathbf{p}\right\}\right)^2 \\ &+ \frac{1}{2}\text{Tr}\left(\Lambda_{i-1}^{-1}\Lambda\right)\text{Tr}\left(\Omega_{i-1}^{-1}\Omega\right) + \frac{1}{2r}\left(\mathbf{p}^\top\Omega\mathbf{p}\right)\left(\mathbf{q}^\top\Lambda\mathbf{q}\right) \ . \end{aligned} \tag{11}$$

The detailed derivation of the update steps is given in a long version [1]. It yields our second algorithm, named f(actored)-AROMA, which is summarized in Fig. 2. Using Woodbury identity it follows that both $\Omega_i$ (13) and $\Lambda_i$ (14) are PSD.

It is worth comparing the update for $\Omega$ (13) in Fig. 2 (S5) with the update of AROW for $\Sigma$ (4). Both updates share the same formal structure, but use different constants. AROW uses the parameter $r$ in the denominator of (4), while f-AROMA uses $mr/\mathbf{q}_i^\top\Lambda_{i-1}\mathbf{q}_i$. Assuming $\|\mathbf{q}_i\|^2 \leq m$ we get that $\frac{mr}{\mathbf{q}_i^\top\Lambda_{i-1}\mathbf{q}_i} \geq r$. Furthermore, the lower the value of $\mathbf{q}_i^\top\Lambda_{i-1}\mathbf{q}_i$ is, the larger is the value of the effective parameter $mr/\mathbf{q}_i^\top\Lambda_{i-1}\mathbf{q}_i$, which in turn reduces the effect of the update. In the extreme case if $\mathbf{q}_i^\top\Lambda_{i-1}\mathbf{q}_i = 0$ then $\Omega_i = \Omega_{i-1}$. Intuitively, the algorithm should decrease the total variance as more examples are observed. Yet, if the variance is already low due to low variance related to the query $\mathbf{q}_i^\top\Lambda_{i-1}\mathbf{q}_i$ then there is no need to reduce the variance related to the output $\Omega$, and vice versa. Following the symmetry between $\Omega$ and $\Lambda$, these observations also hold for the update of $\Lambda$ (14) (S6).

f-AROMA uses a total memory of $mn + m^2 + n^2$ to store the mean matrix $W$ and the covariance matrices $\Omega, \Lambda$. The time complexity is also $mn + m^2 + n^2$ since it involves addition to all elements of these matrices. Note that if $m \approx n$ both d-AROMA and f-AROMA have about the same asymptotic complexity, where the later requires storage and manipulation of one more matrix. When the dimensions $m$ and $n$ differ significantly, $m \ll n$ or $n \ll m$, the complexity of f-AROMA larger than that of d-AROMA because f-AROMA scales quadratically both with $m$ and $n$, while d-AROMA scales linearly with either parameters.

We conclude this section with a mistake bound similar to Theorem 1. Our analysis applies to the algorithm of Fig. 2 with two minor changes. First, it assumes a mistake driven version of the algorithm, namely, that the algorithm makes an update only when a mistake occurs. The condition for an update is therefore $0 > \mathbf{q}_i^\top W_{i-1}\mathbf{p}_i$ instead of $1 > \mathbf{q}_i^\top W_{i-1}\mathbf{p}_i$. Second, from (12) (S4) we get that the update of the factored-AROMA can be written as,

$$\Lambda_{i-1}^{-1}W_i\Omega_{i-1}^{-1} = \Lambda_{i-1}^{-1}W_{i-1}\Omega_{i-1}^{-1} + \frac{\max\left\{0, 1-\mathbf{q}_iW_{i-1}\mathbf{p}_i\right\}}{r + \left(\mathbf{q}_i^\top\Lambda_{i-1}\mathbf{q}_i\right)\left(\mathbf{p}_i^\top\Omega_{i-1}\mathbf{p}_i\right)}\mathbf{q}_i\mathbf{p}_i^\top \ ,$$

the analysis is for a version that uses the new matrices $\Lambda_i$ and $\Omega_i$, that is,

$$\Lambda_i^{-1}W_i\Omega_i^{-1} = \Lambda_{i-1}^{-1}W_{i-1}\Omega_{i-1}^{-1} + \frac{\max\left\{0, 1-\mathbf{q}_iW_{i-1}\mathbf{p}_i\right\}}{r + \left(\mathbf{q}_i^\top\Lambda_{i-1}\mathbf{q}_i\right)\left(\mathbf{p}_i^\top\Omega_{i-1}\mathbf{p}_i\right)}\mathbf{q}_i\mathbf{p}_i^\top \ .$$

We are now ready to state the main theorem of this section.

**Theorem 2** *Let $V$ be any similarity matrix. Assume the algorithm is executed on any sequence of queries and objects, then the total number of mistakes that the algorithm performs is bounded by*

$$|\mathcal{M}| \leq \sum_{i\in\mathcal{M}}\max\left\{0, 1-\mathbf{q}_i^\top V\mathbf{p}_i\right\} + 2\sqrt{\text{Tr}\left(V\Omega_N^{-1}V^\top\Lambda_N^{-1}\right)} \times \sqrt{r\min\{m\log\det\left(\Omega_N^{-1}\right), n\log\det\left(\Lambda_N^{-1}\right)\}} \ .$$

To understand the theorem, the matrices $\Omega_N^{-1}$ and $\Lambda_N^{-1}$ can be thought of as the second order moments of the objects $\mathbf{p}_i$ and the queries $\mathbf{q}_i$ respectively. From (13) (S5) and (14) (S6) we observe that these matrices are the sum of the identity matrix and a weighted sum of outer products of the objects and queries. The first term of the bound

**Algorithm 2: Factored-AROMA**

---

**Input parameters:** A scalar $r$

**Initialize:** $W_0 = 0 \in \mathbb{R}^{m\times n}$, $\Omega_0 = I \in \mathbb{R}^n$, $\Lambda_0 = I \in \mathbb{R}^m$
**For** $i = 1, \ldots, N$

- Sample a query $\mathbf{q}_i \in \mathbb{R}^m$ and two images $\mathbf{p}_i^+, \mathbf{p}_i^- \in \mathbb{R}^n$, such that $similarity(\mathbf{q}_i, \mathbf{p}_i^+) > similarity(\mathbf{q}_i, \mathbf{p}_i^-)$
- Define $\mathbf{p}_i = \mathbf{p}_i^+ - \mathbf{p}_i^-$
- If $1 > \mathbf{q}_i^\top W_{i-1}\mathbf{p}_i$ then update:

$$W_i = W_{i-1} + \frac{\max\left\{0, 1-\mathbf{q}_iW_{i-1}\mathbf{p}_i\right\}}{r + \mathbf{q}_i^\top\Lambda_{i-1}\mathbf{q}_i\mathbf{p}_i^\top\Omega_{i-1}\mathbf{p}_i}\Lambda_{i-1}\mathbf{q}_i\mathbf{p}_i^\top\Omega_{i-1} \tag{12}$$

$$\Omega_i = \Omega_{i-1} - \frac{\mathbf{q}_i^\top\Lambda_{i-1}\mathbf{q}_i}{mr + \left(\mathbf{q}_i^\top\Lambda_{i-1}\mathbf{q}_i\right)\left(\mathbf{p}_i^\top\Omega_{i-1}\mathbf{p}_i\right)}\Omega_{i-1}\mathbf{p}_i\mathbf{p}_i^\top\Omega_{i-1} \tag{13}$$

$$\Lambda_i = \Lambda_{i-1} - \frac{\mathbf{p}_i^\top\Omega_{i-1}\mathbf{p}_i}{nr + \left(\mathbf{p}_i^\top\Omega_{i-1}\mathbf{p}_i\right)\left(\mathbf{q}_i^\top\Lambda_{i-1}\mathbf{q}_i\right)}\Lambda_{i-1}\mathbf{q}_i\mathbf{q}_i^\top\Lambda_{i-1} \tag{14}$$

**Output:** A weight matrix $W_N$ and its confidence $\Sigma_N$

---

*Figure 2.* The f-AROMA algorithm for similarity measures.

$\mathrm{Tr}\left(V\Omega_N^{-1}V^\top\Lambda_N^{-1}\right)$ is small if *either* the rows of $V$ are aligned with eigenvectors of $\Omega_N^{-1}$ associated with small values *or* the columns of $V$ are aligned with eigenvectors of $\Lambda_N^{-1}$, but not necessarily both. This property, (see Sec. 3.1 of Cesa-Bianchi et al., 2005) holds for the input space of AROW, and also for a second order perceptron. For f-AROMA, this property holds for any one of the subspaces, queries or objects.

Next, the second term of the bound is small if either matrices $\Omega_N^{-1}$ and $\Lambda_N^{-1}$ are skewed. This is because the $\log\det$ function is concave. A similar property holds also for Theorem 1 where we required that features from either spaces would be sparse or non-informative. That is, a property is required to hold only for one of the spaces (queries or objects) but not both.

The proof of the theorem relies on the following lemma, which extends Lemma 4 used in the analysis of AROW (Crammer et al., 2009)

**Lemma 3** *The following two bounds hold for the updates in* (13) *(S5) and* (14) *(S6),* $\sum_i (\mathbf{q}_i\Lambda_i\mathbf{q}_i)\left(\mathbf{p}_i^\top\Omega_i\mathbf{p}_i\right) \leq mr\log\det\left(\Omega_N^{-1}\right)$ *and* $\sum_i (\mathbf{q}_i\Lambda_i\mathbf{q}_i)\left(\mathbf{p}_i^\top\Omega_i\mathbf{p}_i\right) \leq nr\log\det\left(\Lambda_N^{-1}\right)$

**Proof:** We prove the first inequality. The second inequality can be proved similarly. Using (14) (S6) we get, $\mathbf{q}_i^\top\Lambda_i\mathbf{q}_i \leq \mathbf{q}_i^\top\Lambda_{i-1}\mathbf{q}_i = mr\frac{\mathbf{q}_i^\top\Lambda_{i-1}\mathbf{q}_i}{mr}$ . Multiplying with $\left(\mathbf{p}_i^\top\Omega_i\mathbf{p}_i\right)$ and summing over $i$ we get,

$$\begin{aligned}\sum_i (\mathbf{q}_i\Lambda_i\mathbf{q}_i)\left(\mathbf{p}_i^\top\Omega_i\mathbf{p}_i\right) &\leq mr\sum_i \frac{\mathbf{q}_i\Lambda_{i-1}\mathbf{q}_i}{mr}\left(\mathbf{p}_i^\top\Omega_i\mathbf{p}_i\right)\\ =mr\sum_i\left(1-\frac{\det\Omega_{i-1}^{-1}}{\det\Omega_i^{-1}}\right) &\leq -mr\sum_i\log\left(\frac{\det\Omega_{i-1}^{-1}}{\det\Omega_i^{-1}}\right)\\ =mr\log\det\Omega_N^{-1},\end{aligned}$$

where the first equality follows from Lemma D.1 of Cesa-Bianchi et al. (2005). ■

**Proof sketch:** (of Theorem 2) We build on previous approach (Orabona & Crammer, 2010) and have the following inequality, which generalizes Corollary 2 of Orabona & Crammer (2010) for matrices.

$$\begin{aligned}|\mathcal{M}| \leq &\sum_{i\in\mathcal{M}}\max\left\{0,1-\mathbf{q}_i^\top V\mathbf{p}_i\right\}+2\sqrt{\mathrm{Tr}\left(V\Omega_N^{-1}V^\top\Lambda_N^{-1}\right)}\\ &\times\sqrt{\sum_{i\in\mathcal{M}}\mathbf{q}_i^\top W_{i-1}\mathbf{p}_i+\sum_{i\in\mathcal{M}}(\mathbf{q}_i\Lambda_i\mathbf{q}_i)\left(\mathbf{p}_i^\top\Omega_i\mathbf{p}_i\right)}\,.\end{aligned}$$

The first sum in the second square-root term is non-positive, as for $i\in\mathcal{M}$ we have $\mathbf{q}_i^\top W_{i-1}\mathbf{p}_i \leq 0$. We use Lemma 3 to bound the second square-root term with, $\sqrt{r\min\{m\log\det\left(\Omega_N^{-1}\right), n\log\det\left(\Lambda_N^{-1}\right)\}}$ , which concludes the proof. ■

## 5. Empirical Evaluation

We evaluated diagonal and factored AROMA on two data sets. First, we learned a semantic similarity between pairs of images in the Caltech-256 dataset (Griffin et al., 2007). Second, we learned a similarity measure between pairs of text documents using the 20-newsgroups data collected by Lang (1995). In both tasks we used standard 5-fold cross validation and report the precision on the test set.

### 5.1. Image similarity in the Caltech256 dataset

We first tested AROMA in an image similarity task using the Caltech256 dataset. This dataset consists of $30,607$ images that were obtained from Google image search and from `PicSearch.com`. Images were assigned to 257 categories and evaluated by humans in order to ensure image quality and relevance. To allow a direct comparisons with the previous literature, we only used here 50 classes.

We represent each image using a sparse code based on a bag of patch descriptors. Specifically, features are extracted by dividing each image into overlapping square patches, and describing each patch with edge and color histograms. For edge histograms, we used *uniform Local Binary Patterns* (uLBPs) (Ojala et al., 2002), which estimate a texture histogram of a patch by considering differences in intensity at circular neighborhoods centered on each pixel. We used uniform $\mathrm{LBP}_{8,2}$ patterns, which means that a circle of radius 2 is considered centered on each block, and bins corresponding to non uniform sequences are merged. LBP patterns were then concatenated with color histograms.

To form a sparse code, patch descriptors were mapped into codewords using a dictionary that was trained over a large set of images using k-means. Then, patch representations were collected to represent an image as a sparse code. Each local descriptor was represented as a discrete index, called *visterm*, and the image was represented as a *bag-of-visterms* vector, in which components $p_i$ are related to the presence or absence of visterm $i$ in $p$. The assignment of the weight $p_i$ of visterm $i$ in image $p$ was according to tf-idf weights. This approach has been found successful (for a related task) by Grangier & Bengio (2008) and Chechik et al. (2009). We used a 1000-sized codebook, with a median of 27 non-zero values per image and a maximum of 129.

We compared the performance of AROMA with five other approaches. **(1) HIER:** *Hierarchical semantic indexing*, an approach that cleverly uses the known hierarchy among class labels (Deng et al., 2011). **(2) OASIS:** An online similarity model based on a ranking cost across triplets, similar to the setup studied here (Chechik et al., 2009). It can be used to estimate the added benefit of using the covariance of the distribution in addition to the mean as AROMA does. **(3) ITML/LEGO** An online approach that succeeds to maintain a proper metric during learning in an efficient way (Davis et al., 2007) **(4) LMNN:** *Large Margin Nearest neighbor*, one of the early large margin metric learning methods (Weinberger et al., 2005). **(5) Euclidean distance:** equivalent to using the identity matrix $W = I$.

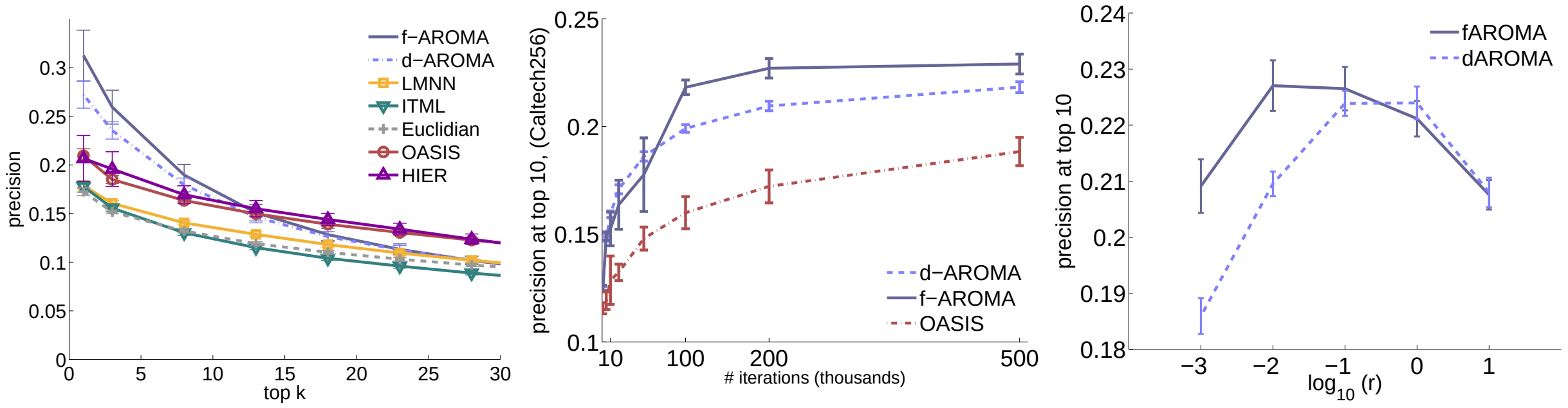

*Figure 3.* Experiments on image similarity using Caltech256. **Left:** Precision as a function of top $k$ images. AROMA was run for 200K iterations, $r = 0.01$. **Middle:** Precision at the 10 top images traced during training. **Right:** Sensitivity of AROMA to the regularizer $r$. Average precision at 10, $200K$ iterations.

The left panel of Fig. 3 compares the precision obtained with d-AROMA and f-AROMA with all other competing methods. Diagonal and factorized AROMA perform very similarly, with a slightly higher performance for factored AROMA. Both methods are significantly better than all other methods at the head of the top ranked images. At the top ranked image, AROMA improves precision by 50% over the second best approach (OASIS, from 22% to 33%).

The middle panel of Fig. 3 traces the precision over the test set during training showing that convergence is achieved after $200K \sim 500K$ iterations. In the beginning d-AROMA was slightly better than f-AROMA, but later f-AROMA converged faster. The right panel of Fig. 3 demonstrates that AROMA is largely robust to the choice of the regularizer $r$, with less than 5% change in precision across three orders of magnitude of $r$.

### 5.2. Document similarity, the 20 Newsgroups dataset

In a second set of experiments we studied the problem of learning a similarity measure between pairs of text documents. This task has numerous applications, such as finding content on the web that is related to a given text document. In this dataset, documents are divided to 20 classes, with about $1,000$ documents in each class. Two documents were considered similar iff they share the same class labels.

We used the 20 newsgroups data set (Lang, 1995) and removed stop words but did not apply stemming. We selected $1,000$ terms that conveyed high information about the identity of the class (over the training set) using the *infogain* criterion (Yang & Pedersen, 1997). The selected features were normalized using *tf-idf*, and then represented each document as a bag of words.

The 20 newsgroups website proposes a split of the data into a train and test sets. We repeated splitting 5 times based on sizes of the proposed splits (a train-to-test ratio of 65% / 35%). We evaluated the learned similarity measures using a ranking criterion. We view every document in the test set $\mathbf{q}$ as a query, and rank the remaining test documents $\mathbf{p}$ by their similarity scores $\mathbf{q}^\top W \mathbf{p}$. We then computed the precision (fraction of positives) at the top $r$ ranked documents. We further computed the *mean average precision* (mAP), a widely used measure in the information retrieval community, which averages over different values of $r$.

With this dataset. we only compared with OASIS and ITML, the methods that achieved higher precision on the Caltech256 data. HIER requires to use a known hierarchy of classes which is not available for the 20NG dataset.

The left panel of Fig. 4 shows the precision at the top ranked similar document. Clearly both AROMA methods outperform ITML and OASIS by large. The middle panel of Fig. 4 traces precision as it progresses through the learning iterations. f-AROMA achieves higher precision than diagonal AROMA during most of the learning iterations, and in fact converges faster. d-AROMA reaches the same level after $500K$ iterations. Interestingly, AROMA learns much faster than OASIS: it takes OASIS ten times more steps to get to the same precision (this effect is also true for the mean average precision). This precision gain is preserved across a large regime of $r$ values, as shown in the right panel of Fig. 4.

## 6. Summary

We presented two algorithms that learn distribution over matrices. Both outperform state-of-the-art methods in two tasks, and model the covariance of the matrix distribution using a linear number of parameters. Diagonal-AROMA is likely to be superior when the variance of individual features is large relative to feature dependencies, and factored-AROMA is expected to be superior when the data has strong correlations across features, as with the Caltech256 data. Factored-AROMA also converged faster.

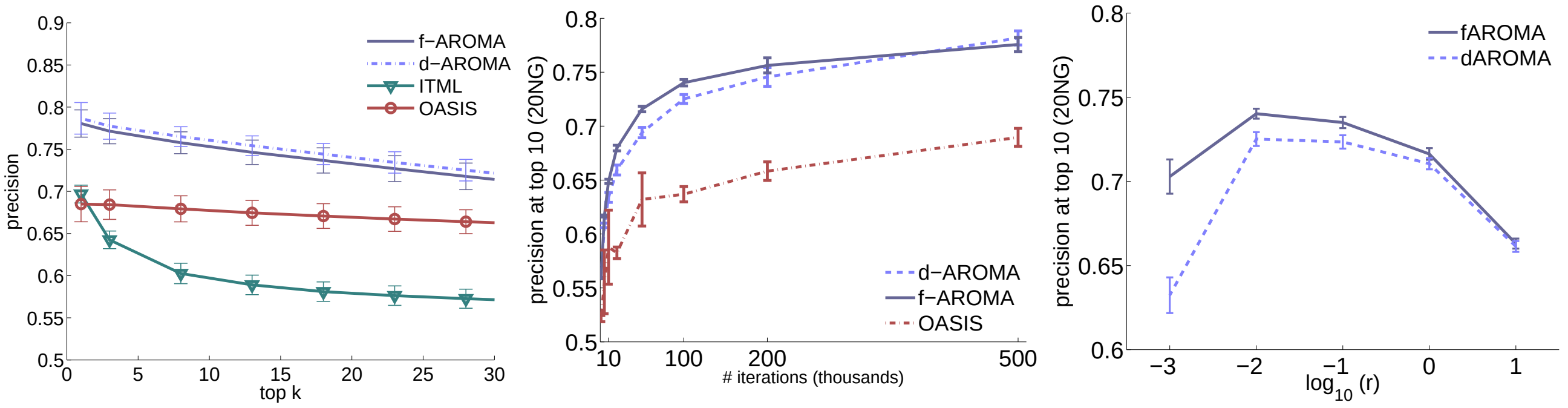

*Figure 4.* 20 Newsgroups. **Left:** Precision at top $k$ of AROMA compared to OASIS and ITML (default parameters). AROMA results were obtained with $r = 1$ and $500K$ iterations. OASIS results were obtained with $c = 0.1$ and $500K$ iterations, and were robust to the choice of $c$. **Middle:** Precision of the 10 top images as a function of training iterations with $r = 0.01$ and a total of $500K$ iterations. **Right:** Sensitivity or AROMA to the regularizer $r$. Average precision over the top 10 images, $100K$ iterations.

**Acknowledgements:** KC gratefully acknowledges partia support by an Israeli Science Foundation grant ISF-1567/10.

## References

Cesa-Bianchi, N., Conconi, A., and Gentile, C. A second-order perceptron algorithm. *Siam Journal of Commutation*, 34(3):640–668, 2005.

Chechik, G., Sharma, V., Shalit, U., and Bengio, S. An online algorithm for large scale image similarity learning. In *NIPS*, 2009.

Crammer, K., Kulesza, A., and Dredze, M. Adaptive regularization of weighted vectors. In *NIPS*, 2009.

Davis, J.V., Kulis, B., Jain, P., Sra, S., and Dhillon, I.S. Information-theoretic metric learning. In *ICML*, 2007.

Deng, J., Berg, A.C., and Fei-Fei, L. Hierarchical semantic indexing for large scale image retrieval. In *cvpr*, 2011.

Dredze, M., Crammer, K., and Pereira, F. Confidence-weighted linear classification. In *ICML*, 2008.

Duchi, J., Hazan, E., and Singer, Y. Adaptive subgradient methods for online learning and stochastic optimization. In *COLT*, pp. 257–269, 2010.

Grangier, D. and Bengio, S. A discriminative kernel-based model to rank images from text queries. *IEEE tran. on pattern analysis and mach. intelligence*, 30(8):1371–1384, 2008.

Griffin, G., Holub, A., and Perona, P. Caltech-256 object category dataset. Technical Report 7694, California Institute of Technology, 2007.

Gupta, A.K. and Nagar, D.K. *Matrix Variate Distributions*. Chapman and Hall/CRC, 1999.

Jain, P., Kulis, B., Dhillon, I., and Grauman, K. Online metric learning and fast similarity search. In *NIPS22*, volume 22, 2008.

Kulis, B., K.Saenko, and T.Darrell. What you saw is not what you get: Domain adaptation using asymmetric kernel transforms. In *CVPR*, pp. 1785–1792, 2011.

Lang, K. Learning to filter netnews. In *ICML*, pp. 331–339, 1995.

McFee, Brian and Lanckriet, Gert. Learning multi-modal similarity. In *JMLR*, volume 12, pp. 491–523, 2012.

Ojala, T., Pietikainen, M., and Maenpaa, T. Multiresolution gray-scale and rotation invariant texture classification with local binary patterns. *IEEE tran. on pattern analysis and mach. intelligence*, 24(7):971–987, 2002.

Orabona, F. and Crammer, K. New adaptive algorithms for online classification. In *NIPS*, 2010.

Weinberger, Kilian Q., Blitzer, John, and Saul, Lawrence K. Distance metric learning for large margin nearest neighbor classification. In *NIPS*, 2005.

Weston, J., Bengio, S., and Usunier, N. Wsabie: Scaling up to large vocabulary image annotation. In *IJCAI*, 2011.

Yang, Y. and Pedersen, J.O. A comparative study on feature selection in text categorization. In *Machine learning-international workshop*, pp. 412–420, 1997.